  \documentclass[twocolumn]{svjour3}         
  \smartqed  
  \usepackage{graphicx}
  
  \usepackage{titlesec} 
  \titleformat{\paragraph}[runin]
  {\bfseries\scshape}{\theparagraph}{1em}{}
  \titlespacing{\paragraph}{0em}{1ex}{.5em} 
  \usepackage{todonotes}
  \usepackage{booktabs}
  \usepackage{tabularx}
  \usepackage{multirow}
  \usepackage{multicol}
  \usepackage{bm}
  \usepackage{url}
  \usepackage{balance}
  \usepackage{enumitem}
  \usepackage{color}
  \usepackage{bbding}
  \usepackage{nccmath}
  \usepackage{amsmath,amssymb}
  \usepackage{graphicx}
  \usepackage{cite}
  
  \usepackage{graphicx}
  \usepackage{amsmath}
  \usepackage{amssymb}
  \usepackage{color}
  \usepackage{epstopdf}
  \usepackage{array}
  \usepackage{multirow}
  \usepackage{amsfonts}
  \usepackage{booktabs}
  \usepackage{float}
  \usepackage{subfigure}
  \usepackage{algorithm}
  \usepackage{algpseudocode}
  \def\redm#1{{\textcolor{black}{#1}}}
  \def\redmrtwo#1{{\textcolor{black}{#1}}}

\begin{document}
  
  \title{StyleAdapter: A Unified Stylized Image Generation Model
  }
  
  
  \author{Zhouxia Wang \and
          Xintao Wang \and
          Liangbin Xie \and
          Zhongang Qi \and \\
          Ying Shan \and
          Wenping Wang \and
          Ping Luo
  }
  
  
  \institute{Zhouxia Wang \at
               The University of Hong Kong, Hong Kong SAR, China \\
                \email{wzhoux@connect.hku.hk}
             \and
             Xintao Wang \at
             ARC Lab, Tencent PCG, Shenzhen, China \\
                \email{xintao.alpha@gmail.com}
             \and
            Liangbin Xie \at
              The University of Macau, Macau SAR, China and Shenzhen Institute of Advanced Technology, Shenzhen, China \\
                \email{lb.xie@siat.ac.cn}
             \and
            Zhongang Qi \at
             ARC Lab, Tencent PCG, Shenzhen, China \\
                \email{zhongangqi@tencent.com}
             \and
            Ying Shan \at
             ARC Lab, Tencent PCG, Shenzhen, China \\
                \email{yingsshan@tencent.com}
             \and
            Wenping Wang \at
             The University of Hong Kong, Hong Kong SAR, China \\
                \email{wenping@cs.hku.hk}
              \and
            Ping Luo \at
             The University of Hong Kong, Hong Kong SAR, China and Shanghai AI Laboratory, Shanghai, China \\
                \email{pluo@cs.hku.hk}
              \and
            Xintao Wang and Ping Luo are the corresponding authors.
  }

  \date{Received: date / Accepted: date}
  
  \maketitle
  

  \def\eg{\emph{e.g.}}
  \def\etal{\emph{et al}}
  
  \newcommand{\EQREF}{Eq.~\eqref}
  \newcommand{\EQSREF}{Eqs.~\eqref}
  \newcommand{\FIGREF}{Fig.~\ref}
  \def\proposed{VB} 
  \def\fixcolor{black}
  \def\hstate{\bm {\tilde s}}
  \def\rstate{\bm s}
  \def\jstate{\bm s^{jn}}
  \def\jpolicy{\overrightarrow{\pi}}
  \def\vpref{v_{\text {pref}}}
  \def\vector#1{\mbox{\boldmath $#1$}}
  \def\sup#1{^{(\rm #1)}}
  \def\sub#1{_{\rm #1}}
  \def\supi#1{^{(#1)}}
  \def\vct#1{\mbox{\boldmath $#1$}}
  \def\eg{{\it e.g.}}
  \def\cf{{\it c.f.}}
  \def\ie{{\it i.e.}}
  \def\etal{{\it et al. }}
  \def\etc{{\it etc}}
  \newcommand{\argmax}{\mathop{\rm argmax}\limits}
  \newcommand{\argmin}{\mathop{\rm argmin}\limits}
  
  \def\Rerr{\Delta \bm r}
  \def\Terr{\Delta \bm t}
  \def\Xerr{\Delta \bm x}
  \def\XerrRel{\Delta \bm {\tilde x}}
  \def\Xgt{\dot{\bm x}}
  \def\Rgt{\dot{R}}
  \def\Tgt{\dot{\bm t}}
  \def\arraystretchlen{1.0}
  
  \def\cam{c}
  \def\image{\mathcal I}
  \def\traj{\mathcal X}
  \def\btraj{\mathcal {\bm X}}
  \def\keypoints{\mathcal P}
  \def\states{\mathcal S}
  \def\bstates{\mathcal {\bm S}}
  \def\state{\bm s}
  \def\ped{\bm x}
  \def\pedi{\bm p} 
  \def\obs{\bm z}
  
  \def\Fi{\bm F_r}
  \def\Fp{\bm F_p}
  \def\vpref{\bm w}
  \def\ENERGY{{\mathcal E}}
  
  \def\DIFF#1{\textcolor{black}{#1}}
  \def\DIFFCR#1{\textcolor{black}{#1}}

  \begin{abstract}
  This work focuses on generating high-quality images with specific style of reference images and content of provided textual descriptions. Current leading algorithms, i.e., DreamBooth and LoRA, require fine-tuning for each style, leading to time-consuming and computationally expensive processes. In this work, we propose StyleAdapter, a unified stylized image generation model capable of producing a variety of stylized images that match both the content of a given prompt and the style of reference images, without the need for per-style fine-tuning. It introduces a two-path cross-attention (TPCA) module to separately process style information and textual prompt, which cooperate with a semantic suppressing vision model (SSVM) to suppress the semantic content of style images. In this way, it can ensure that the prompt maintains control over the content of the generated images, while also mitigating the negative impact of semantic information in style references. This results in the content of the generated image adhering to the prompt, and its style aligning with the style references. Besides, our StyleAdapter can be integrated with existing controllable synthesis methods, such as T2I-adapter and ControlNet, to attain a more controllable and stable generation process. Extensive experiments demonstrate the superiority of our method over previous works.
  \keywords{Stylized Image Generation; Artificial Intelligence Generated Content (AIGC); Diffusion Model; Computer Vision}
  \end{abstract}
  
  \begin{figure*}[ht]
    \centering
    \begin{minipage}[t]{\linewidth}
      \centering
      \includegraphics[width=1\columnwidth]{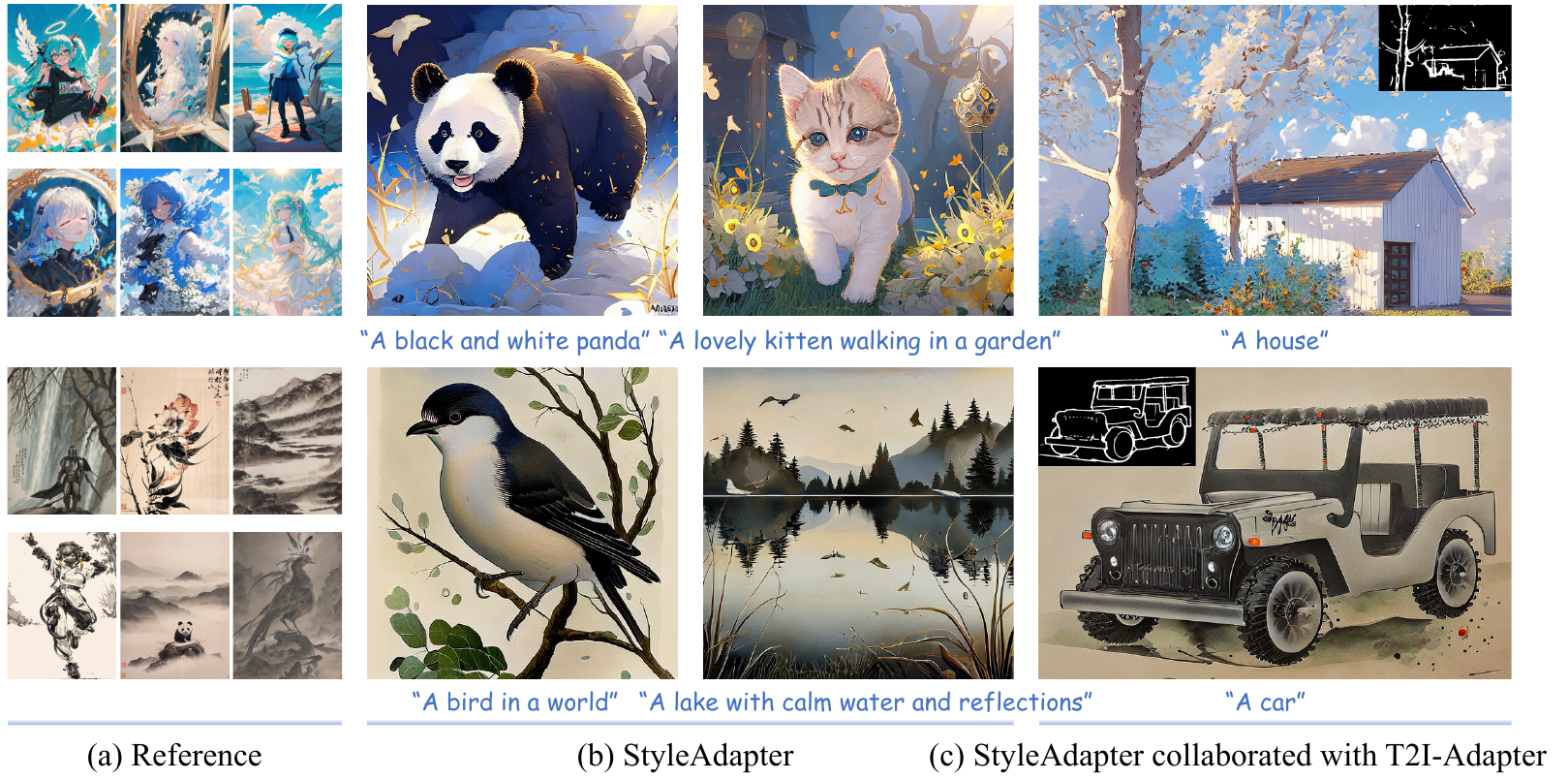}
    \end{minipage}
    \centering
    \caption{Given multiple style reference images, our \textbf{StyleAdapter} is capable of generating images that adhere to both style and prompts without the need for per-style fine-tuning. The style learned from the references primarily focuses on \textbf{brushstrokes, textures, and drawing materials}, such as the specific lines in the first style and the ink material in the second style. Besides, our method shows compatibility with additional controllable conditions, such as sketches.}
    \label{fig:teaser}
  \end{figure*}
  
  \section{Introduction} 
  Recent developments in data and large-scale models have greatly pushed forward the progress in text-to-image (T2I) generation~\cite{t2i3,glid,t2i5,t2i2,ldm,t2i1,t2i4}. These advanced models are skilled in creating high-quality images from given prompts.
  Furthermore, T2I methods can incorporate specific styles into the generated images by using textual descriptions of the style as prompts. However, textual descriptions often lack expressiveness and informativeness compared to visual representations of styles, resulting in T2I outputs with coarse and less detailed style features.
  To leverage the rich information present in visual data of styles, previous works~\cite{TI,zhang2022inversion} have proposed textual inversion methods that map visual representations of styles to textual space. This approach enables the style information extracted from visual images to guide T2I models. Nevertheless, these methods still face limitations, as the visual-to-textual projection fails to preserve the rich details inherent in visual images, leading to suboptimal styles in the generated images.
  Currently, DreamBooth~\cite{dreambooth} and LoRA~\cite{lora} offer more effective solutions by employing fine-tuning to the original diffusion model or utilizing extra small networks to adapt to specific styles. These approaches enable the generation of images with relatively precise styles, capturing details such as brushstrokes and textures. However, the need to fine-tune or re-train the model for each new style makes these methods computationally demanding and time-consuming, rendering them impractical for many applications.
  
  Developing a unified model capable of generating various stylized images without per-style fine-tuning is highly desirable for increased efficiency and flexibility. This work aims to propose such a unified model to generate high-quality stylized images that match the content of a given prompt and the style of the style references. However, accurately extracting style information from style images and ensuring that the style information and textual prompts precisely focus on stylization and content generation, respectively, remains a significant challenge. Our vanilla approach reveals that simply extracting style reference features with the vision model of CLIP~\cite{clip} and combining them with prompt features as the condition for Stable Diffusion (SD)~\cite{ldm} leads to two main issues: \textit{1) loss of prompt controllability over generated content, and 2) inheritance of both semantic and style features from style references, compromising content fidelity.}
  
  Our in-depth observations and analyses demonstrate that separately injecting contextual prompt and semantic-suppressed style reference information into generated images can effectively ensure prompt controllability and mitigate the negative impact of semantic information in style references. Based on these analyses, we propose StyleAdapter, a unified stylized image generation model that produces a variety of stylized images matching both the content of a given prompt and the style of reference images without per-style fine-tuning. It introduces a two-path cross-attention (TPCA) module to separately process style information and textual prompts, cooperating with a semantic suppressing vision model (SSVM) to suppress style image semantics. This ensures prompt controllability over generated content while mitigating the negative impact of semantic information in style references. Furthermore, StyleAdapter can be integrated with existing controllable synthesis methods, such as T2I-adapter~\cite{mou2023t2i} and ControlNet~\cite{controlnet}, for a more controllable and stable generation process.

  Our contributions can be summarized as follows:
  \begin{itemize}
      \item We propose StyleAdapter, a unified stylized image generation model capable of producing a variety of stylized images that match both the content of a given prompt and the style of reference images, without requiring per-style fine-tuning.
      \item Based on in-depth observations and analyses, we introduce a two-path cross-attention (TPCA) module to separately process style information and textual prompts. By further cooperating with a semantic suppressing vision model (SSVM) for suppressing the semantic content of style images, it ensures the controllability of the prompt over the generated content while mitigating the negative impact of semantic information in style references.
      \item Our StyleAdapter can be integrated with existing controllable synthesis methods to generate high-quality images in a more controllable and stable manner.
  \end{itemize}
  
  The remainder of this paper is structured as follows: Section~\ref{sec:related_work} begins with a review of existing literature related to text-to-image synthesis and stylized image generation. This is followed by Section~\ref{sec:method}, where we delve into a detailed analysis of the challenges faced by the Vanilla StyleAdapter. In this section, we also introduce our proposed StyleAdapter, designed to overcome these challenges. The effectiveness of our proposed StyleAdapter is then assessed through a series of experiments and evaluations, as detailed in Section~\ref{sec:exp}. The paper concludes with Section~\ref{sec:conclusion}, where we summarize our findings and contributions.
  
  \section{Related Works}
  \label{sec:related_work}
  
  \subsection{Text-to-image synthesis}
  
  Text-to-image synthesis (T2I) is a challenging and active research area that aims to generate realistic images from natural language text descriptions. Generative adversarial networks (GANs)  are one of the most popular approaches for T2I synthesis, as they can produce high-fidelity images that match the text descriptions~\cite{reed2016generative,zhang2018stackgan++,xu2018attngan,li2019controllable,chen2024dynamic,chen2024heterogeneous}. However, GANs suffer from training instability and mode collapse issues~\cite{brock2018large,dhariwal2021diffusion,ho2022cascaded}. Recently, diffusion models have shown great success in image generation~\cite{song2020denoising,diff,nichol2021improved,dhariwal2021diffusion}, surpassing GANs in fidelity and diversity. Many recent diffusion methods have also focused on the task of T2I generation. For example, Glide~\cite{glid} proposed to incorporate the text feature into transformer blocks in the denoising process. Subsequently, DALL-E~\cite{t2i2}, Cogview~\cite{t2i3}, Make-a-scene~\cite{gafni2022make}, Stable Diffusion~\cite{ldm}, and Imagen~\cite{saharia2021image} significantly improved the performance in T2I generation. To enhance the controllability of the generation results, ControlNet~\cite{controlnet} and T2I-Adapter~\cite{mou2023t2i} have both implemented an additional condition network in conjunction with stable diffusion. This allows for synthesizing images that adhere to the text and condition.
  
  \subsection{Stylized image generation}
  
  
  Image style transfer is a task that involves generating artistic images guided by an input image. Traditional style transfer methods match patches between content and style images using low-level hand-crafted features~\cite{wang2004efficient,zhang2013style}. With the rapid development of deep learning, deep convolutional neural networks have been employed to extract the statistical distribution of features that effectively capture style patterns~\cite{gatys2016image,gatys2017controlling,kolkin2019style}. In addition to CNNs, visual transformers have also been utilized for style transfer tasks~\cite{wu2021styleformer,stytr2}. 
  
  \redm{
  Recently, leveraging the success of diffusion models~\cite{ldm,saharia2021image,t2i2}, the application of these powerful generative models for image stylization has gained significant attention. For instance, InST~\cite{zhang2022inversion} employs diffusion models as a backbone for inversion and as a generator for stylized image creation. Methods such as Textual Inversion~\cite{TI}, P+~\cite{P+}, and ProSpect~\cite{Prospect} map style reference images into the textual embedding space, incorporating their learned textual embeddings into either the entire diffusion model, specific blocks of UNet, or certain denoising steps to guide the generation of stylized images. Despite their achievements in image stylization, these methods face limitations as the visual-to-textual projection struggles to retain the rich details inherent in visual images. 
  Other approaches, such as DreamBooth~\cite{dreambooth}, LoRA~\cite{lora}, and StyleDrop~\cite{sohn2023styledrop}, propose fine-tuning the stable diffusion (SD) model for specific concepts or styles. While effective, these methods require fine-tuning the SD model for each concept or style, thus limiting their generalization. In contrast, our StyleAdapter aims to generate various stylized images with a unified model, eliminating the need for per-style fine-tuning. Instantstyle~\cite{Instantstyle}, a late work, shares a similar concept with our StyleAdapter. Built on Ip-adapter~\cite{Ip-adapter}, it balances text and image prompts through decoupled cross-attention and adapts image stylization by embedding subtraction and manual reference image injection. 
  Distinct from this approach, StyleAdapter suppresses the semantic information of the reference image by thoroughly considering the structure of the feature extractor and the characteristics of the reference images. Additionally, our method employs explicitly layer-wise learned weights to combine Text Cross-Attention and Style Cross-Attention, resulting in more stable performance in generative image stylization.
  }
  
  \section{Methodology}
  \label{sec:method}
  
  This work aims to propose a unified stylized image generation model capable of producing a variety of stylized images that match both the content of a given prompt and the style of reference images, without the need for per-style fine-tuning. This work builds upon SD. In this section, we first briefly describe the SD and the vision model in CLIP~\cite{clip} commonly used to extract vision features from vision data. Then, we introduce a vanilla StyleAdapter, which highlights the challenges in constructing a unified stylized image generation model. Based on in-depth observations and analyses, we propose our delicate StyleAdapter, with a two-path cross-attention (TPCA) module for separately processing style information and textual prompts, and a semantic suppressing vision model (SSVM) for suppressing style image semantics. This approach ensures prompt controllability over generated content while mitigating the negative impact of semantic information in style references.
  
  \subsection{Preliminary}
  \label{sec:preliminary}
  
  \noindent \textbf{Stable Diffusion. }
  Stable Diffusion (SD) is a latent diffusion model (LDM)~\cite{ldm} trained on large-scale data and LDM is a generative model that can synthesize high-quality images from Gaussian noise by iterative sampling. Compared to the traditional diffusion model, its diffusion process happens in the latent space.
  Therefore, except for a diffusion model, an autoencoder consisting of an encoder $\mathcal{E(\cdot)}$ and a decoder $\mathcal{D(\cdot)}$ is needed. $\mathcal{E(\cdot)}$ is used to encode an image $I$ into the latent space $z$ ($z = \mathcal{E}(I)$) while $\mathcal{D(\cdot)}$ is used to decode the feature in the latent space back to an image.
  The diffusion model contains a forward process and a reverse process. Its denoising model $\epsilon_{\theta}(\cdot)$ is implemented with UNet~\cite{UNet} and trained with a simple mean-squared loss:
  \begin{equation}
  L_{L D M}:=\mathbb{E}_{z \sim \mathcal{E}(I), c, \epsilon \sim \mathcal{N}(0,1), t}\left[\left\|\epsilon-\epsilon_{\theta}\left(z_{t}, t, c\right)\right\|_{2}^{2}\right],
  \label{eq:ldm}
  \end{equation}
  where $\epsilon$ is the unscaled noise, $t$ is the sampling step, $z_t$ is latent noise at step $t$, and $c$ is the condition. 
  While SD acts as a T2I model, $c$ is the text feature $f_t$ of a natural language prompt extracted with the text model of CLIP~\cite{clip}.
  $f_t$ is then integrated into SD with a cross-attention model, whose query $\mathbf{Q}_t$ is from the spatial feature $y$ which is extracted from $Z_t$, and key $\mathbf{K}_t$ and value $\mathbf{V}_t$ are from $f_t$.
  The process can be expressed as:
  \begin{equation}
  \left \{
  \begin{array}{ll}
      \mathbf{Q}_t = \mathbf{W}_{Qt} \cdot y;\ \mathbf{K}_t = \mathbf{W}_{Kt}\cdot f_t;\ \mathbf{V}_t = \mathbf{W}_{Vt}\cdot f_t; \\
      Attention(\mathbf{Q}_t, \mathbf{K}_t, \mathbf{V}_t) = softmax(\frac{\mathbf{Q}_t\mathbf{K}_t^T}{\sqrt{d}})\cdot \mathbf{V}_t,
  \end{array}
  \right.
  \label{eq:ldm_ca}
  \end{equation}
  where $\mathbf{W}_{Q_t/K_t/V_t}$ are learnable weights, and $d$ is dependent on the number of channels of $y$.
  
  \noindent\textbf{Vision Model of CLIP.}
  The vision model in CLIP~\cite{clip} is commonly used for extracting features from vision data in T2I models. To process a vision image, such as our style reference $I_r\in\mathbb{R}^{(H \times W \times C)}$ ($H, W, C$ are the hight, width, and channels, respectively), the vision model takes a sequence of its flattened patches $I_r^p\in\mathbb{R}^{N \times (P^2\cdot C)}$ ($P$ is the patch size and $N=HW/P^2$ is the sequence length) as input, and deploys a vision embedding module to attain their embeddings with a linear projection $\mathbf{E}\in\mathbb{R}^{P^2\cdot C\times D}$. An additive class embedding $E_{cls}\in\mathbb{R}^{1\times D}$ is attached to the vision embeddings before adding with position embedding $E_{pos}\in\mathbb{R}^{(N+1)\times D}$. The embedding process can be formulated as:
  \begin{equation}
      E_{Ir} = [E_{cls}, I_r^0\mathbf{E}, I_r^1\mathbf{E}, I_r^{N-1}\mathbf{E}] + E_{pos}.
      \label{eq:vit}
  \end{equation}
  Then the $E_{Ir}$ is encoded into vision features $f_r$ with a vision encoder.

  \begin{figure}
      \centering
      \includegraphics[width=0.3\textwidth]{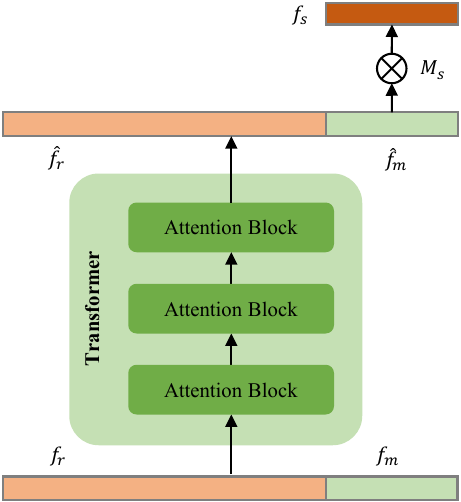}
      \caption{\textbf{Structure of StyleEmb.} StyleEmb contains a learnable embedding $f_m$. After concatenating it with style feature $f_r$, it updates $f_m$ with a transformed network by extracting useful information from $f_r$. Its updated embedding $\hat{f}_m$ is then mapped to $f_s$ with a learnable matrix $M_s$.}
      \label{fig:StyleEmb}
  \end{figure}
  
  \begin{figure*}[t]
  \setlength\tabcolsep{1pt}
  \scriptsize
  \centering
  \begin{minipage}[t]{\linewidth}
  \centering
  \includegraphics[width=0.95\columnwidth]{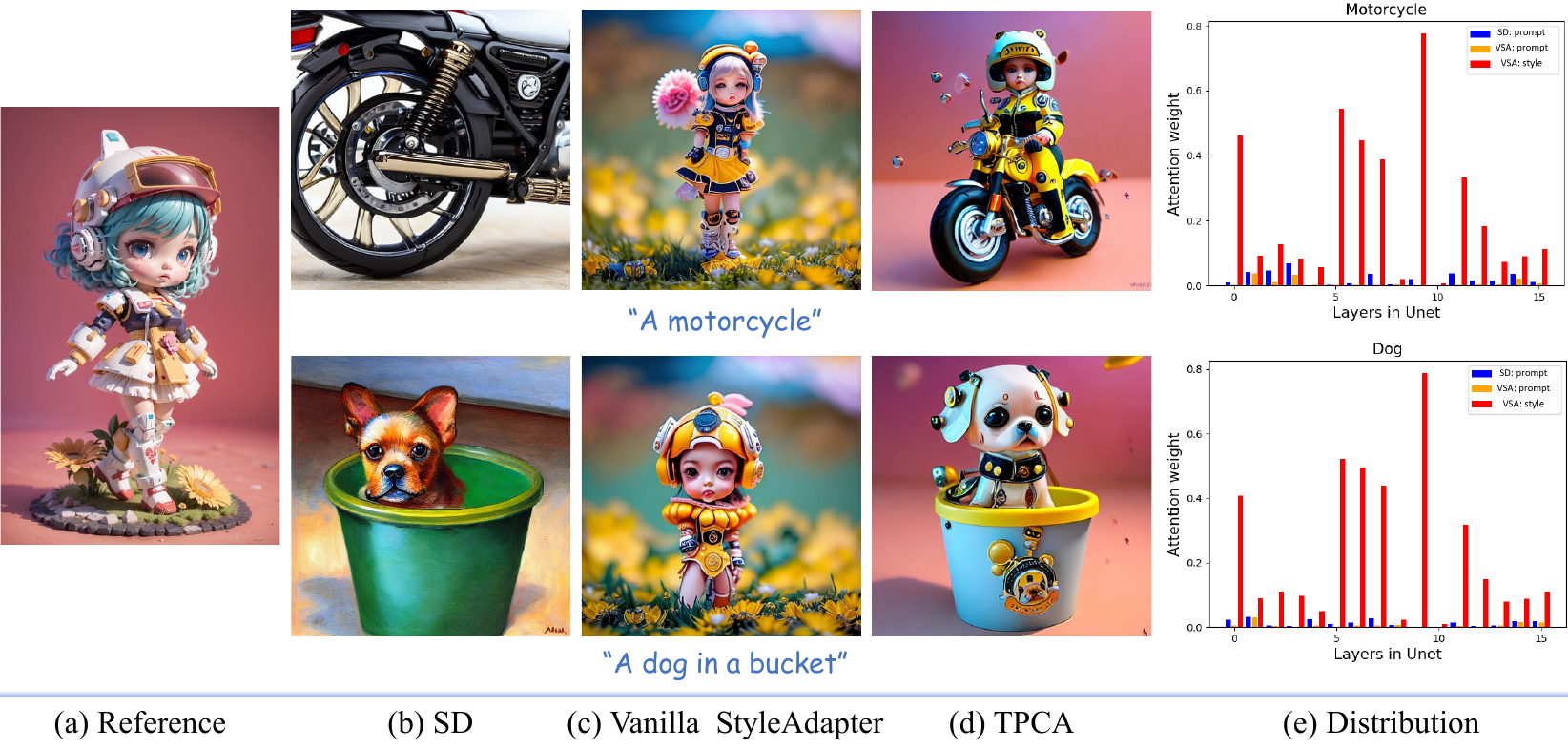}
  \end{minipage}
  \centering
  \caption{
  \textbf{Illustration of prompt controllability loss.}
  Without style reference, SD~\cite{ldm} generates images matching content prompts, such as the motorcycle and dog in (b). However, \textcolor{black}{Vanilla StyleAdapter (VSA)} concatenates style reference features with prompts, resulting in images dominated by the girl and flowers in the style image, as shown in (c). (e) is the attention weights of keywords (motorcycle and dog) in SD and \textcolor{black}{VSA}, which reveal that after combining prompt with style features, \textcolor{black}{VSA} reduces prompt attention and focuses more on style features. We propose a two-path cross-attention module (TPCA) to inject prompt and style reference features into the generated images separately, preserving both content and style, as shown in (d).}
  \label{fig:dominant}
  \end{figure*}

  \begin{figure*}[t]
  \setlength\tabcolsep{1pt}
  \scriptsize
  \centering
  \begin{minipage}[t]{\linewidth}
  \centering
  \includegraphics[width=0.95\columnwidth]{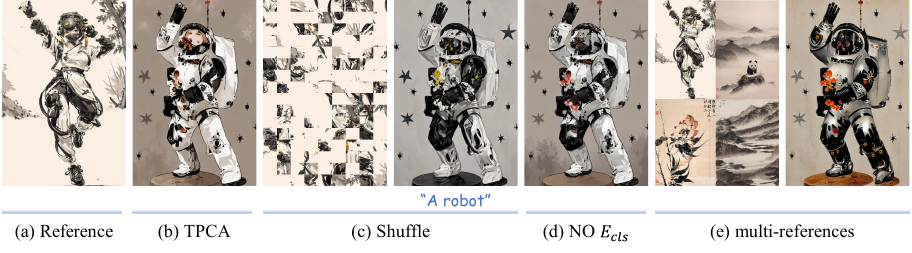}
  \end{minipage}
  \centering
  \caption{\textbf{Preliminary experimental results on the issue of semantic and style coupling in the style image}. (b) shows a result of our TPCA. It is a robot whose style is similar to the reference but with a human face, due to the tight coupling between the semantic and style information in the reference. Our preliminary experimental results in (c)-(e) respectively showcase that patch-wisely shuffling the reference image, removing the class embedding $E_{cls}$ in Eq.~\ref{eq:vit}, and providing multiple diverse reference images can help mitigate this issue.}
  \label{fig:semantic}
  \end{figure*}
  
  \begin{figure*}[!t]
      \centering
      \includegraphics[width=0.95\linewidth]{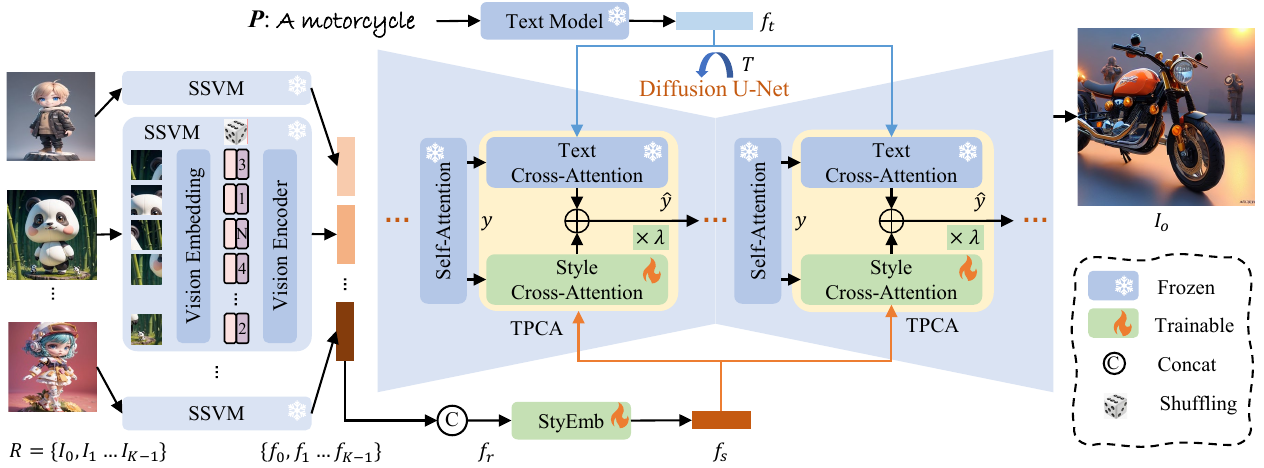}
      \caption{\textbf{StyleAdapter Framework.} StyleAdapter is built upon SD~\cite{ldm} and utilizes CLIP's~\cite{clip} text model to extract the features of prompt $P$. It employs a semantic suppressing vision model (SSVM) to extract style information from multiple style reference $R$, and suppresses their semantic information by shuffling the patch-based vision embeddings and removing the original class embedding. Then, the reference features are concatenated as $f_r$ and processed by the StyEmb Module to obtain style feature $f_s$. The prompt feature $f_t$ and style feature $f_s$ are separately processed using the two-path cross-attention module (TPCA) before fusing with learnable coefficient $\lambda$. The fused result is passed to the subsequent SD block. After $T$ sampling steps, StyleAdapter generates a stylized image with content matching the prompt and style conforming to the references.}
      \label{fig:framework}
  \end{figure*}

  \subsection{Vanilla StyleAdapter with In-Depth Analyses}
  
  A straightforward approach to adapt SD to stylized image generation involves extracting the style feature $f_r$ from a style reference $I_r$ using the vision model of CLIP~\cite{clip} and concatenating it with the prompt feature ($f_t$). This concatenated result serves as the condition for guiding the generation of SD.
  To enhance the expressiveness of the style feature, we employ an additive style embedding module (\textbf{StyEmb}) with a transformer block to embed $f_r$ into $f_s$. As illustrated in Fig.~\ref{fig:StyleEmb}, StyEmb predefines a learnable embedding $f_m$ which is appended to $f_r$. After processing with a transformer consisting of three attention blocks, we attain $\hat{f}_m$, which is further projected to the style feature $f_s$ using a learnable matrix $M_s$. Note that $f_m$ extracts information from $f_r$ and can adapt to $f_r$ with a flexible length (referring to our later multiple references). By concatenating $f_t$ and $f_s$ as the condition $c$ in Eq.~\ref{eq:ldm} ($c = [f_t, f_s]$), we can generate stylized images with SD.

  As shown in Fig.~\ref{fig:dominant} (c), this vanilla approach can achieve a desirable stylization effect. However, it reveals two major challenges:
  1) \textit{
  the prompt loses controllability over the generated content,
  }
  and 2) 
  \textit{the generated image inherits both the semantic and style features of the style reference images, compromising its content fidelity.
  }
  
  By further analyzing the results of the original SD and the vanilla StyleAdapter, we get an observation.
  
  
  \noindent\textbf{Observation 1: Simply combining the features of the prompt and style reference potentially results in a loss of prompt controllability over the generated content.}
  Fig.~\ref{fig:dominant} (b) shows that the original SD generates natural images confirming the prompt content, e.g., the motorcycle and dog. However, when adapting it to stylized image generation implemented with the vanilla StyleAdapter, the prompts lose their controllability over the generated content, and the content in the style reference becomes dominant. As shown in (c), the girl from (a) becomes the main object. We explore the insight reason by plotting the attention weights of "motorcycle", "dog", and style features in each cross-attention layer of SD or vanilla StyleAdapter. Statistic results in (e) reveal that the attention to "motorcycle" and "dog" decreases when involving style features to guide the image generation, while style features gain higher attention. This suggests that simply combining the features of the prompt and style reference makes it difficult to properly utilize these two information sources during stylized image generation. To address this issue, we employ a two-path cross-attention module (TPCA, detailed in~\ref{sec:TPCA}) to process these sources separately. Corresponding results in (d) demonstrate that prompts regain controllability over the generated content.
  %

  Nonetheless, the second challenge remains unresolved. Results in Fig.~\ref{fig:dominant} (d) and Fig.~\ref{fig:semantic} (b) indicate that both the content specified in the prompt and style reference appears in the generated images, such as the robot body and natural human face in Fig.~\ref{fig:semantic} (b). This issue primarily stems from the tight coupling between semantic and style information in the style reference, leading to another insightful observation.
  
  \noindent\textbf{Observation 2: Semantic suppressing is required when extracting style features from style references.}
  Considering that the vision model described in Section~\ref{sec:preliminary} extracts style features patch-wisely and its class embedding $E_{cls}$ has been proven to be rich in semantic information for classification~\cite{vit}, we aim to shuffle these patches and remove $E_{cls}$ to disrupt and reduce the semantic information in style references. Corresponding generated results are in Fig.~\ref{fig:semantic}, which successfully replace the natural human face with a robot face. Moreover, the result in (e) suggests that using multiple style images with diverse semantics (e.g., human, panda, flower, and mountain) and similar styles (e.g., ink style) enables the generation model to extract similar style information and disregard their diverse semantic information. These phenomena inspire us to propose a \textbf{semantic suppressing vision model} (SSVM) with multiple style references to obtain semantic-suppressed style features for stylized image generation.

  \subsection{StyleAdapter}
  \label{sec:styleadapter}
  
  Motivated by previous observations and analyses, we propose our delicate StyleAdapter, deployed with a two-path cross-attention module (TPCA) to separately process style information and textural prompt, and cooperated with a semantic suppressing vision model (SSVM) to suppress the semantic content in style images.
  
  Specifically, as depicted in Fig.~\ref{fig:framework}, our StyleAdapter is based on SD, with conditions comprising a natural language prompt $\mathbf{P}$ and style reference images $\mathbf{R}=\{I_0, I_1, \dots, I_{K-1}\}$. The textual feature $f_t$ is extracted using a traditional text model~\cite{clip}, while the style features $\{f_0, f_1, \dots, f_{K-1}\}$ are extracted using our proposed SSVM. These style features are processed into $f_s$ using the style embedding module (\textbf{StyEmb}). Subsequently, $f_t$ and $f_s$ are independently incorporated into the generation process using our proposed TPCA, before being combined with a learnable weight $\lambda$. The fused result is passed to the subsequent SD block. After $T$ sampling steps, we generate image $I_o$, conforming to the desired content and style. StyleAdapter is learned with $L_{LDM}$ (Eq.~\ref{eq:ldm}), where condition $c$ consists of $f_t$ and $f_s$.

  \subsubsection{Tow-Path Cross-Attention Module}
  \label{sec:TPCA}
  
  We deploy our two-path cross-attention module after each self-attention module in the diffusion U-Net~\cite{UNet} model. It consists of two parallel cross-attention modules: the Text Cross-Attention module and the Style Cross-Attention module, which are responsible for handling the prompt-based condition and the style-based condition, respectively. The query of both cross-attention modules comes from the spatial feature $y$ of SD. However, the key and value of the Text Cross-Attention come from the text feature $f_t$, while the key and value of the Style Cross-Attention come from the style feature $f_s$. 
  The attention output of Text Cross-Attention $Attention(\mathbf{Q}_t, \mathbf{K}_t, \mathbf{V}_t)$ has the same formula as Eq.~\ref{eq:ldm_ca}, while the output of Style Cross-Attention \\ $Attention(\mathbf{Q}_s, \mathbf{K}_s, \mathbf{V}_s)$ is formulated as:
  \begin{equation}
  \left \{
  \begin{array}{ll}
      \mathbf{Q}_s = \mathbf{W}_{Qs} \cdot y;\ \mathbf{K}_s = \mathbf{W}_{Ks}\cdot f_s;\ \mathbf{V}_s = \mathbf{W}_{Vs}\cdot f_s; \\
      Attention(\mathbf{Q}_s, \mathbf{K}_s, \mathbf{V}_s) = softmax(\frac{\mathbf{Q}_s\mathbf{K}_s^T}{\sqrt{d}})\cdot \mathbf{V}_s.
  \end{array}
  \right.
  \end{equation}
  %
  The outputs of these two attention modules are then added back to $y$ and fused with a learnable parameter $\lambda$. This produces a new spatial feature $\hat{y}$ that is fed to the subsequent blocks of SD. The process can be expressed as:
  \begin{equation}
      \hat{y} = Attention(\mathbf{Q}_t, \mathbf{K}_t, \mathbf{V}_t) + \lambda  Attention(\mathbf{Q}_s, \mathbf{K}_s, \mathbf{V}_s).
  \label{eq:tpca}
  \end{equation}
  It is worth noting that since SD already has a strong representation of the prompt, we retain the original cross-attention in SD as our Text Cross-Attention and freeze it during training. In contrast, we train the Style Cross-Attention module to endow it with the capability of fusing style features from the references to the generated images. 
  
  \subsubsection{Semantic Suppressing Vision Model}
  \label{sec:SSVM}
  Our semantic suppressing vision model (SSVM) aims to suppress the semantic information in style references while extracting style features, mitigating the negative impact on the generated images. It achieves semantic suppressing from three aspects.
  i) It removes $E_{cls}$ in Eq.~\ref{eq:vit}, which is rich in semantic information. ii) It patch-wisely shuffles the reference image by randomly shuffling the $E_{pos}$ in Eq.~\ref{eq:vit} before adding them to the patch embeddings. iii) It adopts multiple semantic-diverse style images as references. 
  
  Specifically, SSVM parallelly processes multiple style references $R = \{I_0, I_1, \dots, I_{K-1}\}$, where $K$ denotes the number of style reference images. $K = 3$ while training, although it can be any positive integer. It patch-wisely extracts the vision embedding with a vision embedding module, and directly adds the vision embedding with randomly shuffled position embedding ($E_{pos}$). Then added results are sent into a vision encoder to attain the vision features of each reference, which will be further processed with StyEmb and Style Cross-Attention modules.
  
  
  \begin{figure*}[t]
  \centering
  \begin{minipage}[t]{\linewidth}
  \centering
  \includegraphics[width=0.95\columnwidth]{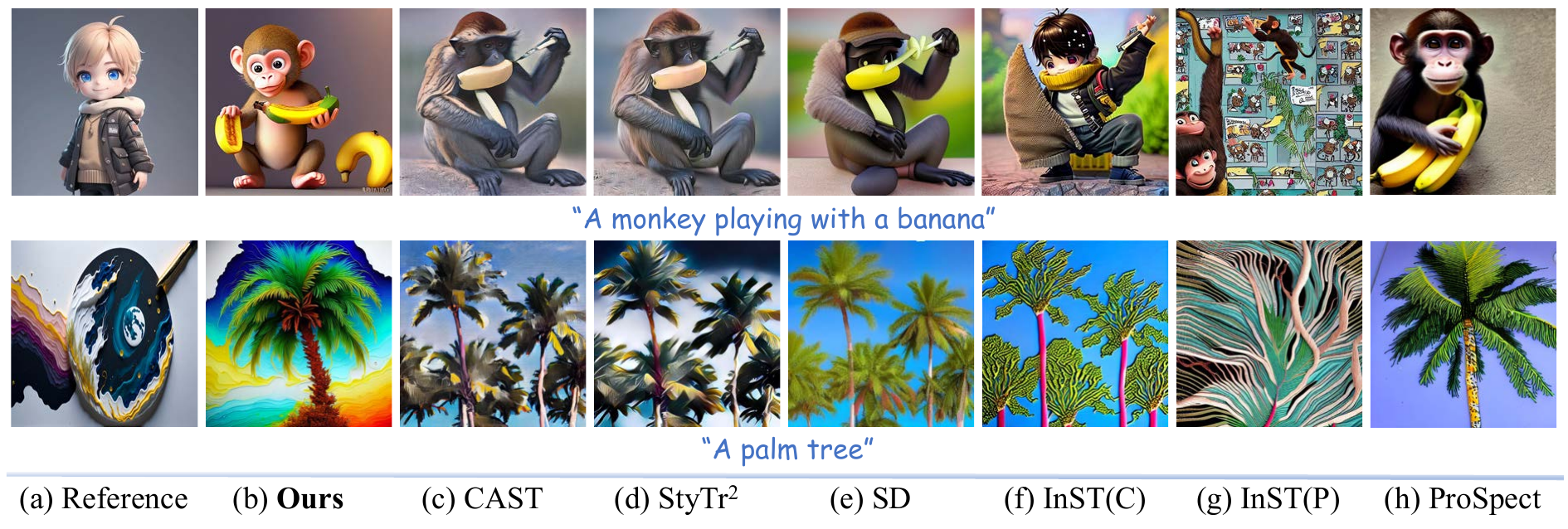}
  \end{minipage}
  \centering
  \caption{\redm{Qualitative comparison with state-of-the-art methods using a single style reference image: Traditional methods like CAST~\cite{cast} and StyTr$^2$~\cite{stytr2} focus on color transfer, whereas diffusion-based methods like SD~\cite{ldm} and InST~\cite{InST} struggle with content-style balance. Our StyleAdapter captures more style details from references, such as \textbf{brushstrokes and textures}, while better matching prompt content.}}
  \vspace{-3pt}
  \label{fig:single_examplar_comparsion} 
  \end{figure*}

  
  \begin{figure*}[t]
  \centering
  \begin{minipage}[t]{\linewidth}
  \centering
  \includegraphics[width=0.95\columnwidth]{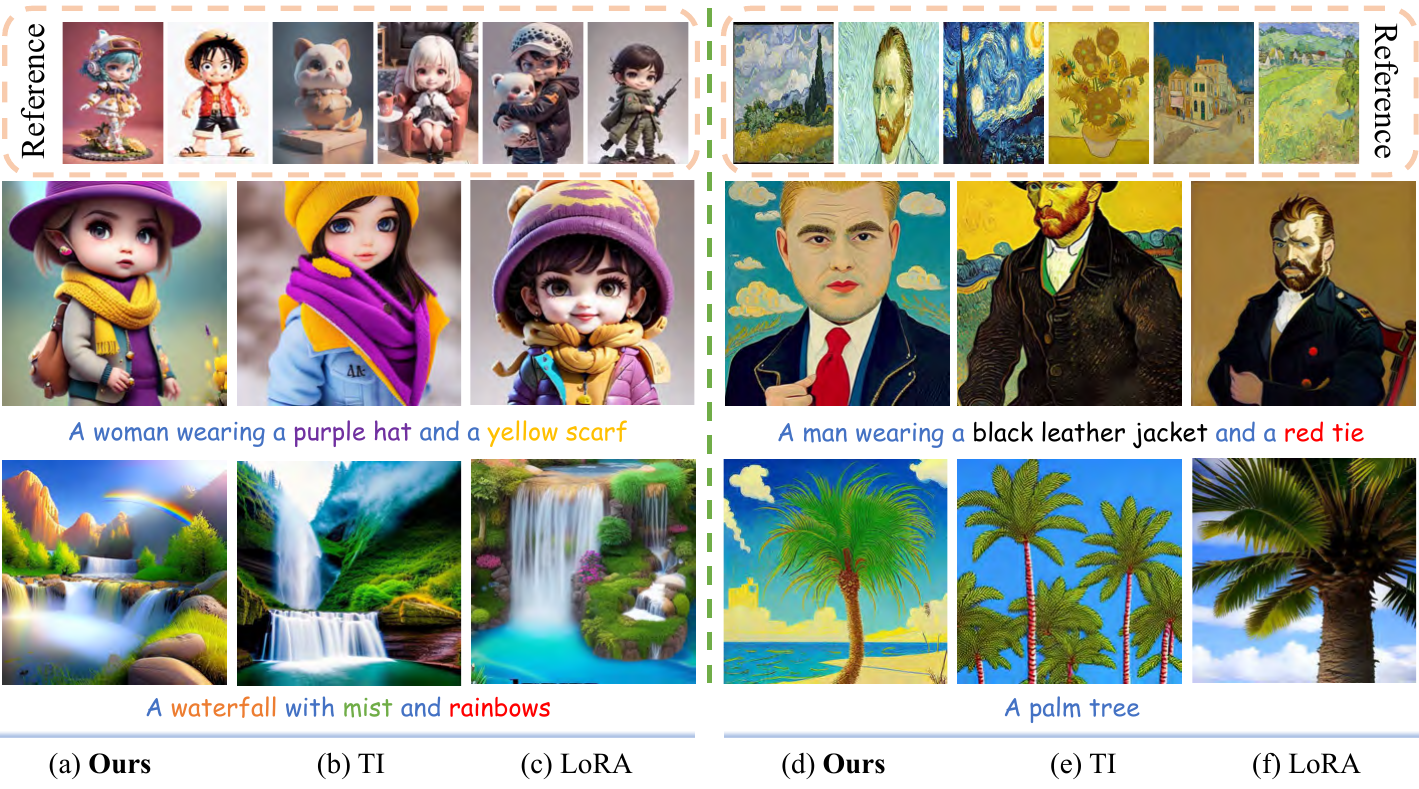}
  \end{minipage}
  \centering
  \caption{Qualitative comparison with TI~\cite{TI} and LoRA~\cite{lora} using multiple style reference images. TI and LoRA, trained on references, perform well in stylization but show insensitivity to the prompt. Conversely, our StyleAdapter, without requiring per-style fine-tuning, performs better in generating both style and content.}
  \vspace{-10pt}
  \label{fig:multiple_examplar_comparison} 
  \end{figure*}
  
  \section{Experiments}
  \label{sec:exp}
  
  \subsection{Experimental settings}
  
  \noindent \textbf{Datasets.} 
  We employ a subset of the \\ LAION-AESTHETICS~\cite{schuhmann2022laion} dataset, containing 600K image-text pairs, for training. While training, the style references are the augmentation results attained from the image in the text-image pairs in the \\ LAION-AESTHETICS.
  
  
  To evaluate the effectiveness of our proposed method compared with previous related methods, we construct a diverse testset that consists of 50 prompts, 50 content images, and 8 groups of style references, leading to a total of 400 evaluation pairs. 
  Although our proposed method does not require content images, we still provide them to meet the requirement of the content-based methods, such as CAST~\cite{cast} and StyTR$^2$~\cite{stytr2}. These content images are generated with SD with prompts in the testset, which are used to determine the content of the generated images of the prompt-based methods, including our proposed StyleAdapter.
  Furthermore, the style references are collected from the Internet\footnote{The style references are collected from https://civitai.com, https://wall.alphacoders.com, and https://foreverclassicgames.com. }. Each style contains 5 to 14 images. On one hand, they are used as the reference for the reference-based stylized method, such as CAST~\cite{cast}, StyTR$^2$~\cite{stytr2}, and our StyleAdapter. On the other hand, they are used as training data for finetuning-based methods, such as Texture Inversion~\cite{TI} and LoRA~\cite{lora}.
  More details are in the supplementary materials.
  
  \begin{table*}[t]
  \renewcommand{\arraystretch}{1.3}
  \centering
  \caption{\textbf{Objective quantitative comparisons with the state-of-the-art methods}. Our StyelAdapter achieves a better balance in text similarly, style similarity, and quality.}
    \begin{tabular}{c|ccccccc|ccc} 
      \hline
       & \multicolumn{7}{c|}{\textbf{Single-reference}} & \multicolumn{3}{c}{\textbf{Multi-reference}} \\
      \hline
      Methods & CAST & StyTr$^2$ & InST(P) & InST(C) & SD &  \redm{ProSpect} & \textbf{Ours} & TI & LoRA & \textbf{Ours} \\
      \hline
      Text-Sim $\uparrow$ & 0.2323 & 0.2340 & 0.2204 & 0.1682 & 0.2145 &\redm{0.2464}  & 0.2435 & 0.1492 & 0.2390 & 0.2448 \\  
      Style-Sim $\uparrow$ & 0.8517 & 0.8493 & 0.8616 & 0.8707 & 0.8528 & \redm{0.8620} & 0.8645  & 0.9289 & 0.9034 & 0.9031 \\
      FID $\downarrow$ & 163.77 & 151.45 & 177.91 & 153.45 & 189.34 & \redm{160.66} & 141.78 & 139.56 & 137.40 & 140.97 \\
      \hline
    \end{tabular}
  \label{tab:sota}
  \end{table*}
  
  \noindent \textbf{Implementation Details.} 
  %
  %
  Our StyleAdapter is deployed on SD model~\cite{ldm} in version 1.5 and CLIP~\cite{clip} implemented with a large ViT~\cite{vit} (patch size 14). The original SD model and blocks from CLIP are frozen during training. Only the StyEmb and the Style Cross-Attention module proposed in this work are trainable. They are optimized with Adam~\cite{adam} optimizer. The learning rate is $8\times10^{-6}$ while the batch size is 8. Experiments run on 8 NVIDIA Tesla 32G-V100 GPUs. Input and style images are resized to $512 \times 512$ and $224 \times 224$, respectively. 
  \redmrtwo{During training, we synthesize style reference images by augmenting the images in the training data through operations such as random cropping, resizing, horizontal flipping, and rotation. These augmentations help prevent shortcut learning while providing style information, such as brushstrokes. We use $K=3$ style references during training, with $K$ being variable during the inference phase.}
  We set sampling step $T=50$ for inference.
  
  \noindent \textbf{Evaluation metrics.}
  %
  This paper evaluates generated images both \textbf{subjectively} and \textbf{objectively} in terms of text similarity, style similarity, and quality. We conduct a \textbf{User Study} for subjective assessment and employ a CLIP-based~\cite{clip} metric to objectively measure text similarity (Text-Sim) and style similarity (Style-Sim) using cosine similarity.
  \redmrtwo{Specifically, we extract the visual features of the generated images, the style references, and the text features from the prompt using ViT-L/14~\cite{ViT-L-14}. We then calculate the cosine similarity between the visual features of the generated images and the style references to determine the Style-Sim score, and between the visual features of the generated images and the text features from the prompt to derive the Text-Sim score.}
  Additionally, we utilize FID~\cite{fid} to assess image quality.

  \subsection{Comparisons with State-of-the-art Methods}
  \redm{
  In this section, we conduct comparisons with current state-of-the-art methods, including two traditional style transfer methods: CAST~\cite{cast} and StyTr$^2$, and four SD-based methods: InST~\cite{InST}, Textual Inversion (TI)~\cite{TI}, LoRA~\cite{lora}, SD~\cite{ldm}, and ProSpect~\cite{Prospect}.
  Specifically, for SD~\cite{ldm}, we utilize its image-to-image mode to generate a stylized image from the content image in the test set, using prompts generated from the reference image with BLIP2~\cite{blip2}. For ProSpect~\cite{Prospect}, the learned token embedding that represents the reference image is involved only in the $6_{th}$ to $10_{th}$ denoising stages, which are claimed to function on stylization.
  }
  
  \subsubsection{Comparisons based on single style reference.}
  We compare our method with state-of-the-art methods, implemented with single style reference, in Fig.~\ref{fig:single_examplar_comparsion}. While CAST~\cite{cast} and StyTr$^2$~\cite{stytr2} perform relatively coarse-grained color transfer, SD~\cite{ldm} yields unsatisfactory stylization due to the poor text representation from reference images. InST~\cite{InST}, based on textural inversion, can generate stylized images using content image (InST(C)) or prompt (InST(P)). InST(C) outperforms previous methods and InST(P) in stylization, but its content is dominated by the style reference image or its generated texture is unnatural. For instance, it generates a boy rather than the monkey indicated in the prompt in the first sample, and the twisted texture in the second sample leads to a strange appearance of the generated image. Besides, InST(P) generates content closer to the prompt but with different styles.
  \redm{
  In comparison, a more recent work, ProSepct~\cite{Prospect}, demonstrates better performance in generated content and style, particularly in content consistency, as evidenced by a higher Text-Sim score in Table~\ref{tab:sota}. This is because the style reference information is involved only in the $6_{th}$ to $10_{th}$ denoising stages, thereby not affecting the generation of the structure, which is typically established in the earlier denoising stages. However, this also leads to a limitation: its effectiveness in stylization is less pronounced compared to our StyleAdapter. In our approach, the reference information is incorporated throughout the entire denoising stage, with minimal negative impact on content consistency due to our delicately designed two-path cross-attention module cooperating with semantic suppression vision model.
  }

  \begin{table}[t]
  \renewcommand{\arraystretch}{1.3}
  \centering
  \caption{\textbf{Subjective quantitative comparisons with the state-of-the-art methods}. Our StyelAdapter attains more preference from expert users.}
    \begin{tabular}{ccccc} 
      \hline
      Methods & CAST & InST(P) & LoRA & \textbf{Ours} \\
      \hline
      Text-Sim $\uparrow$ & 0.2310 & 0.0548 & 0.2869 & \textbf{ 0.4274}\\
      Style-Sim $\uparrow$ & 0.3857 &  0.0286 & 0.1881 & \textbf{0.3976}  \\
      Quality $\uparrow$ & 0.2071 & 0.0452 & 0.3238 & \textbf{0.4238} \\
      \hline
    \end{tabular}
  \label{tab:user_study}
  \end{table}
  
  \begin{figure*}[t]
  \centering
  \begin{minipage}[t]{\linewidth}
  \centering
  \includegraphics[width=0.95\columnwidth]{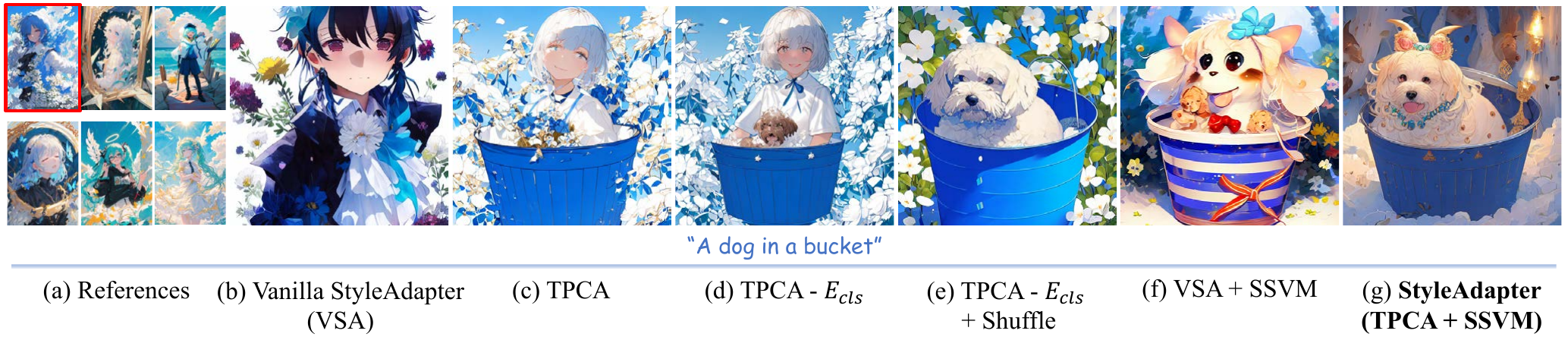}
  \end{minipage}
  \centering
  \caption{\textbf{Qualitative results of ablation studies on TPCA and SSVM.} Results (b)$\sim$(e) are attained with the \textbf{single} reference in the \textcolor{red}{red box}. (b) is the result attained with the vanilla StyleAdapter (VSA), whose content and style are both dominated by the reference image. When updating the vanilla StyleAdapter with our proposed Two-Path Cross-Attention (TPCA) module, the `bucket' in the prompt appears in (c), but the girl in the style reference still exists. By further removing the class embedding $E_{cls}$ in the vision model used for style feature extracting, the `dog' appears in (d). Further shuffling the style reference totally suppresses the girl from the reference image and the prompt dominates the generated content. However, the style of (e) is a little bit far away from the style reference. (g) is attained with the full setting of StyleAdapter. By taking all the images in the same style in (a) as references, the result in (g) not only keeps the dominance of prompt over the generated content but achieves the style from the reference. It is important to note that the color palette in (g) is extracted from all the reference images collectively, rather than from a specific individual reference image. \redm{Additionally, we present result (f), which is obtained by directly applying SSVM to the Vanilla StyleAdapter. However, its performance falls short of the results achieved in (g) in terms of both content consistency and stylization quality. \textbf{The detailed setting of each experiment is corresponding to Table~\ref{tab:abs}.}}}
  \label{fig:ablation} 
  \end{figure*}
  
  \subsubsection{Comparisons based on multiple style reference.}
  Unlike our method, which is a unified model that can be generalized to different styles without per-style fine-tuning, TI~\cite{TI} and LoRA~\cite{lora} require training on the style reference images for each style. Fig.~\ref{fig:multiple_examplar_comparison} and Table~\ref{tab:sota} present the qualitative and quantitative results, respectively. TI~\cite{TI} inverses the style references into a learnable textural embedding embedded into the prompt for guiding the generation. It performs better in style similarity (high score in Style-Sim). However, Its generated content cannot match the prompt accurately, such as the purple hat, yellow scarf, red tie, and rainbows, indicated in the prompts but missed in their corresponding generated results, leading to a lower score in Style-Text. Our StyleAdapter is comparable to LoRA~\cite{lora} in style, but it performs better in text similarity, according to the higher score of Text-Sim and the generated tie and rainbows responding to the prompts in the visualized results, which demonstrates that our StyleAdapter achieves a better balance between content similarity, style similarity, and generated quality in objective metrics.
  
  \subsubsection{User Study.} 
  To attain a more comprehensive evaluation, we conduct a user study. We randomly select 35 generated results covering all styles and employ 24 users who worked in AIGC to evaluate these generated results in three aspects: text similarity, style similarity, and quality. Consequently, we received a total of 2520 votes, and the results in Table~\ref{tab:user_study} show that the generated results obtained with our StyleAdapter are more preferred in all three aspects. We observe that the difference between objective and subjective metrics mainly lies in the fact that objective evaluation metrics independently assess each aspect, while users may take the information from the other aspects into consideration even though we provide separate options. This demonstrates that our generated results achieve a better trade-off between quality, text similarity, and style similarity.

  \subsubsection{Cooperation with existing adapters.} 
\begin{figure*}[th]
  \centering
  \begin{minipage}[t]{\linewidth}
  \centering
  \includegraphics[width=0.85\columnwidth]{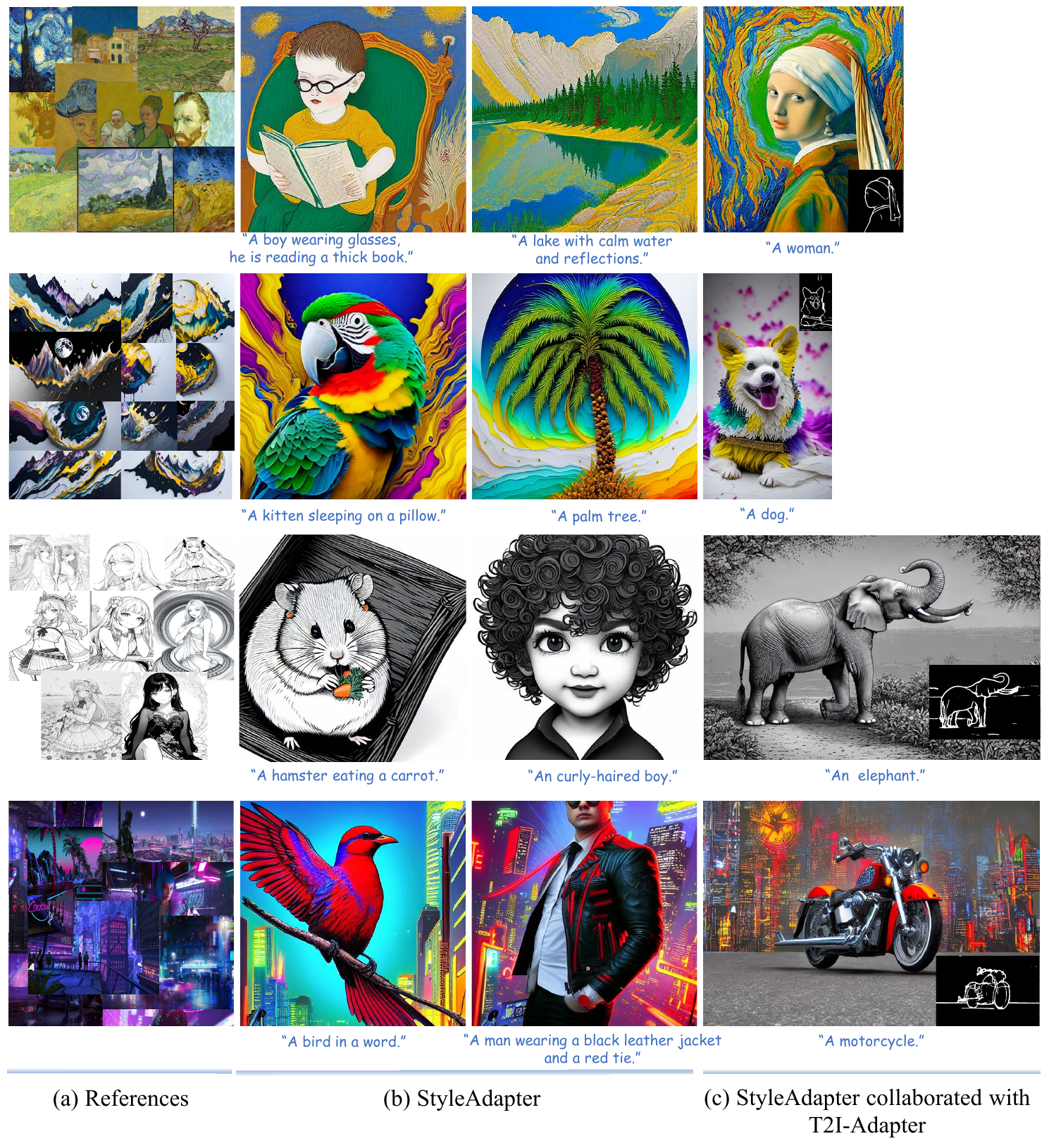}
  \end{minipage}
  \centering
  \caption{
  \textbf{More generated results.} Given multiple style reference images, our StyleAdapter can generate images that adhere to both style and prompts in a single pass. Moreover, our method shows compatibility with additional controllable conditions, such as sketches.
  }
  \label{fig:more_results} 
  \end{figure*}

\begin{figure*}[th]
  \centering
  \begin{minipage}[t]{\linewidth}
  \centering
  \includegraphics[width=0.85\columnwidth]{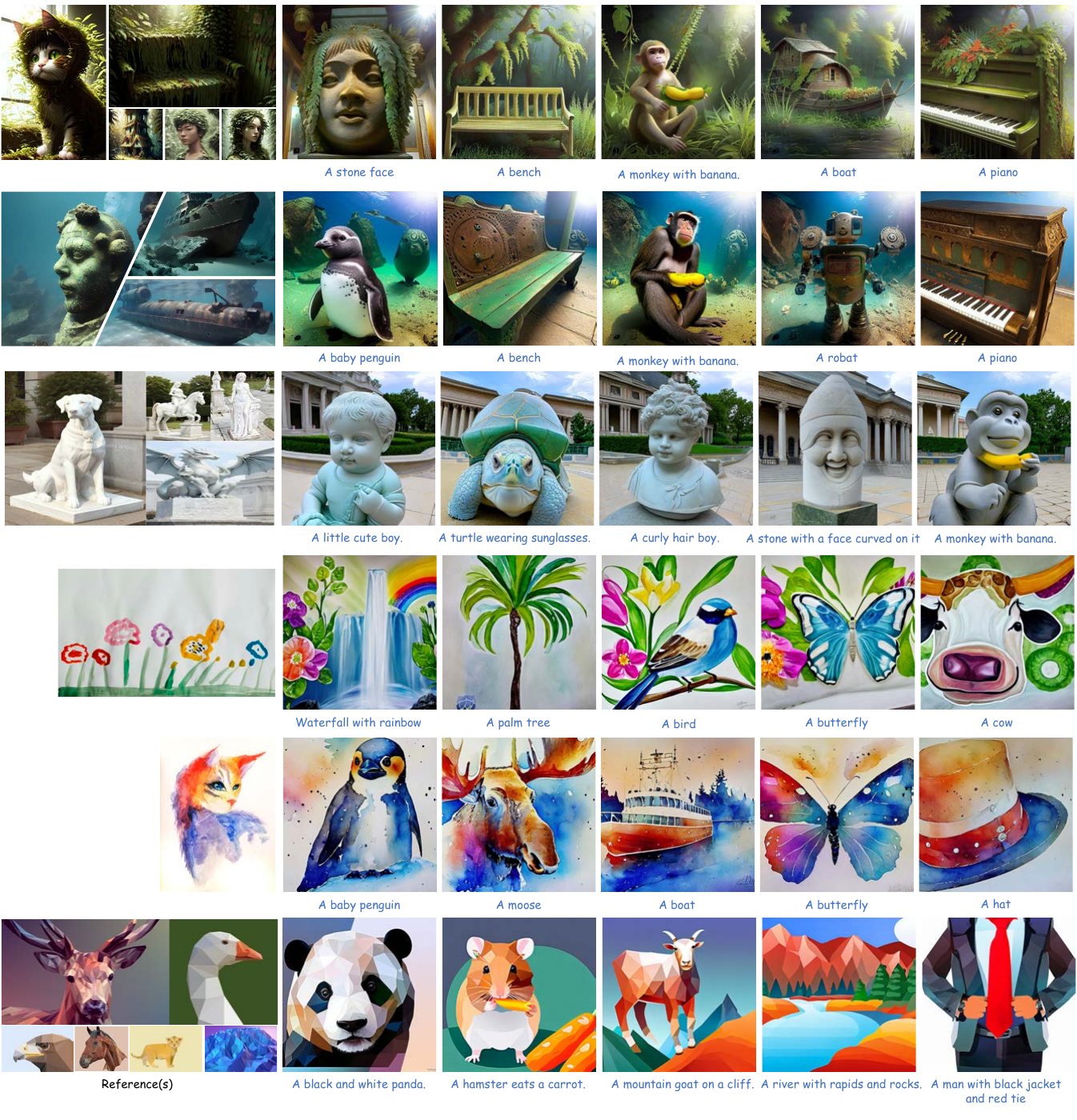}
  \end{minipage}
  \centering
  \caption{
  \redm{
  \textbf{More generated results.} Given multiple style reference images, our StyleAdapter can generate images that adhere to both style and prompts in a single pass. The styles from top to down are objects covered with leaves and laying in the forest, objects under the sea, objects carved with stone, kid's drawing, watercolor style, and origami style.
  }
  }
  \label{fig:more_results_v1} 
  \end{figure*}

  Our StyleAdapter can cooperate with existing adapters, such as T2I-adapter~\cite{mou2023t2i}. Results are in the last column of Fig.~\ref{fig:teaser} and more visualized results of our proposed StyleAdapter in Fig.~\ref{fig:more_results} and Fig.~\ref{fig:more_results_v1}. It shows that with the guidance of the additional sketches, the shape of the generated contents is more controllable, while the style of the generated results still conforms to their corresponding style references.

  
  \begin{table*}[th]
  \caption{\textbf{Quantitative results of ablation studies on TPCA and SSVM.} Our method based on TPCA achieves a significant improvement in Text-Sim compared to Vanilla StyleAdapter (VSA). Employing strategies in SSVM can progressively improve Text-Sim, and eventually attain a better balance between Text-Sim and Style-Sim after utilizing multiple references. Moreover, our TPCA and SSVM enhance the quality of generated images, as indicated by the lower FID score. \redm{We also provide the quantitative results of experiment (f), which directly applies SSVM on Vanilla StyleAdapter. Although it can improve the score of Text-Sim, it falls behind the comprehensive StyleAdapter configuration in both Text-Sim and Style-Sim.}}
  \centering
    \begin{tabular}{c|c|c|c|c|c|c|c|c} 
      \hline
      Exp. & \textcolor{black}{VSA} & TPCA & No $E_{cls}$ & Shuffling & multi-reference & Text-Sim~$\uparrow$ & Style-Sim~$\uparrow$ & FID$~\downarrow$ \\
      \hline
      (b) VSA &  \checkmark & &           &            &                  &  0.1263  & 0.9362  & 186.17 \\
      (c) TPCA & &  \checkmark &           &            &                &  0.2089  & 0.8963 & 145.37 \\
      (d) TPCA - $E_{cls}$ & & \checkmark & \checkmark &            &               &  0.2109  & 0.8921 & 141.99 \\
      (e) TPCA - $E_{cls}$ + suffle & & \checkmark & \checkmark & \checkmark  &              &  0.2435  & 0.8645 & 141.78 \\
      \redm{(f) VSA + SSVM} & \redm{\checkmark} & & \redm{\checkmark} & \redm{\checkmark}  &  \redm{\checkmark}  &  \redm{0.2411}  & \redm{0.8955} & \redm{160.82} \\
      (g) \textbf{StyleAdapter} & & \checkmark & \checkmark & \checkmark  &  \checkmark  &  \textbf{0.2448}  & \textbf{0.9031} & \textbf{140.97} \\
      \hline
    \end{tabular}
    \label{tab:abs}
  \end{table*}

  \subsection{Ablation Studies}
  \label{sec:ablation}
  
  \subsubsection{Effectiveness of TPCA and SSVM}
  
  We evaluate the effectiveness of our proposed Two-Path Cross-Attention (TPCA) module and Semantic Suppressing Vision Model (SSVM) through experiments, with qualitative results in Fig.~\ref{fig:ablation}. Using Vanilla \\StyleAdapter to fuse the prompt and single style reference information (the reference in \textcolor{red}{red} box), as in (b), the content is dominated by the reference girl, ignoring the prompt's dog and bucket. To improve prompt controllability, we process the prompt and style reference separately using our proposed TPCA module, resulting in (c). It shows that the bucket appears but the dog is missing and the girl in the reference image remains. This issue stems from the tight coupling between semantic and style information in the style reference. We remove the class embedding $E_{cls}$ via our SSVM. The result in (d) shows that the dog and bucket appear but it still includes the girl from the style reference. Further shuffling the patches in SSVM, as in (e), disrupts the tight coupling between the semantic and style information in the style reference, removing the reference girl and emphasizing the prompt's dog in the bucket, but with a less similar style to the reference. We further employ multiple style references and attain result (g). It showcases that its content is dominated by the prompt and style closely resembling the reference images. These outcomes demonstrate the effectiveness of our TPCA module and SSVM. 
  
  We also provide corresponding quantitative results, which are in Table~\ref{tab:abs}. It showcases that compared to Vanilla StyleAdapter, our StyleAdapter that uses TPCA achieves higher scores in terms of Text-Sim, which means its results are more consistent with the prompts, although sacrificing some performance of stylization (as indicated by the lower score in terms of Style-Sim). To further suppress the semantic information in the style references while extracting style information, we employ a semantic suppressing vision model (SSVM) to extract style features. By removing $E_{cls}$, SSVM can slightly improvement of the score of Text-Sim while barely affecting the performance of stylization. Further adopting patch-wise shuffling significantly suppresses the semantic information in the style references and boosts the score of Text-Sim by about 0.0326. However, it also degrades the style of the generated results considerably, as shown by the large drop in the score of Style-Sim. By further taking multiple references as input, our StyleAdapter enhances both Text-Sim and Style-Sim, achieving a better balance between the content and style of the generated results. Moreover, our TPCA and SSVM enhance the quality of generated images, as indicated by the lower FID score.
  
  \begin{table}[t]
  \renewcommand{\arraystretch}{1.3}
  \centering
  \caption{
  \redmrtwo{\textbf{Quantitative results of StyleEmb.} Compared to methods that combine multiple style references by averaging and concatenating, our final design— which combines multiple references using a learnable embedding ($f_m$) —achieves superior performance in both content and style consistency.}}
    \begin{tabular}{cccc} 
      \hline
      Methods & Average & ConCat & \textbf{Ours} \\
      \hline
      Text-Sim $\uparrow$ & 0.2161 & 0.2094 & \textbf{ 0.2448}\\
      Style-Sim $\uparrow$ & 0.8951 &  0.8959 & \textbf{0.9031}  \\
      \hline
    \end{tabular}
  \label{tab:ab_styleemb}
  \end{table}
  
  \begin{figure}[ht]
  \centering
  \begin{minipage}[t]{\linewidth}
  \centering
  \includegraphics[width=1\columnwidth]{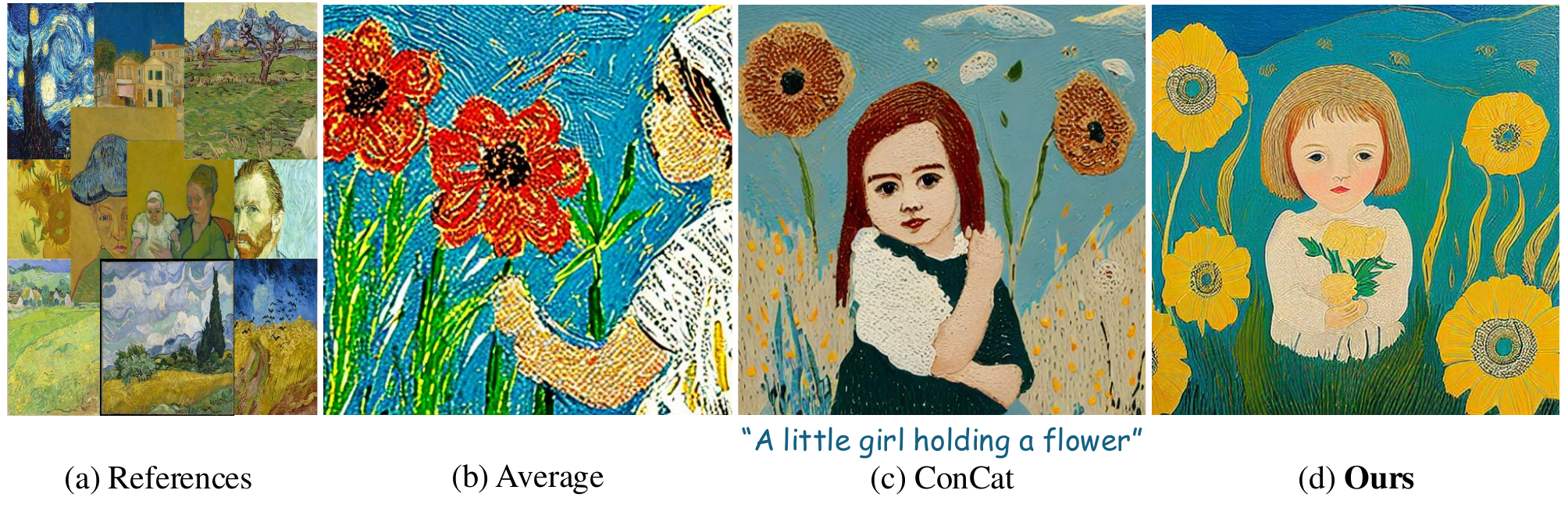}
  \end{minipage}
  \centering
  \caption{
  \redmrtwo{\textbf{Qualitative results of StyleEmb.} Compared to methods that combine multiple style references by averaging and concatenating, our final design— which combines multiple references using a learnable embedding ($f_m$) —achieves superior performance in both content and style consistency.}}
  \label{fig:ab_styleemb} 
  \end{figure}
  
  \redm{
  To conduct a comprehensive evaluation, we further explore the impact of eliminating the TPCA and directly applying SSVM to the Vanilla StyleAdapter. 
  The qualitative and quantitative results, as depicted in Fig.~\ref{fig:ablation} and Table~\ref{tab:abs} respectively, reveal that suppressing the semantics of style references individually can enhance the controllability of the text prompt over the generated content. However, this approach falls short of the performance of StyleAdapter (exp (g)), as evidenced by the less faithful dog image in Fig.~\ref{fig:ablation} (f) and the lower Text-Sim score in Table~\ref{tab:abs}.
  That is mainly due to the mechanism of combination between text prompt and style reference information. Vanilla StyleAdapter combines these two types of features before injecting into the cross-attention module, which inherently determines which information receives greater attention, resulting in the potential dismissal of the text prompt feature, even when efforts have been made to suppress its semantic influence. In contrast, our TPCA proposes to inject these two types of information individually with two separate cross-attentions before weightedly combining them. Importantly, the weight assigned to the text prompt feature is fixed at 1, while the weight for style images is learned during training (as described in Eq.~\ref{eq:tpca}). This design effectively maintains controllability of text prompt over the generated content while allowing for flexible stylization.
  Therefore, StyleAdapter, combining TPCA with SSVM, achieves superior performance in both content and style consistency.
  }
  
  \begin{figure*}[ht]
  \centering
  \begin{minipage}[t]{\linewidth}
  \centering
  \includegraphics[width=0.9\columnwidth]{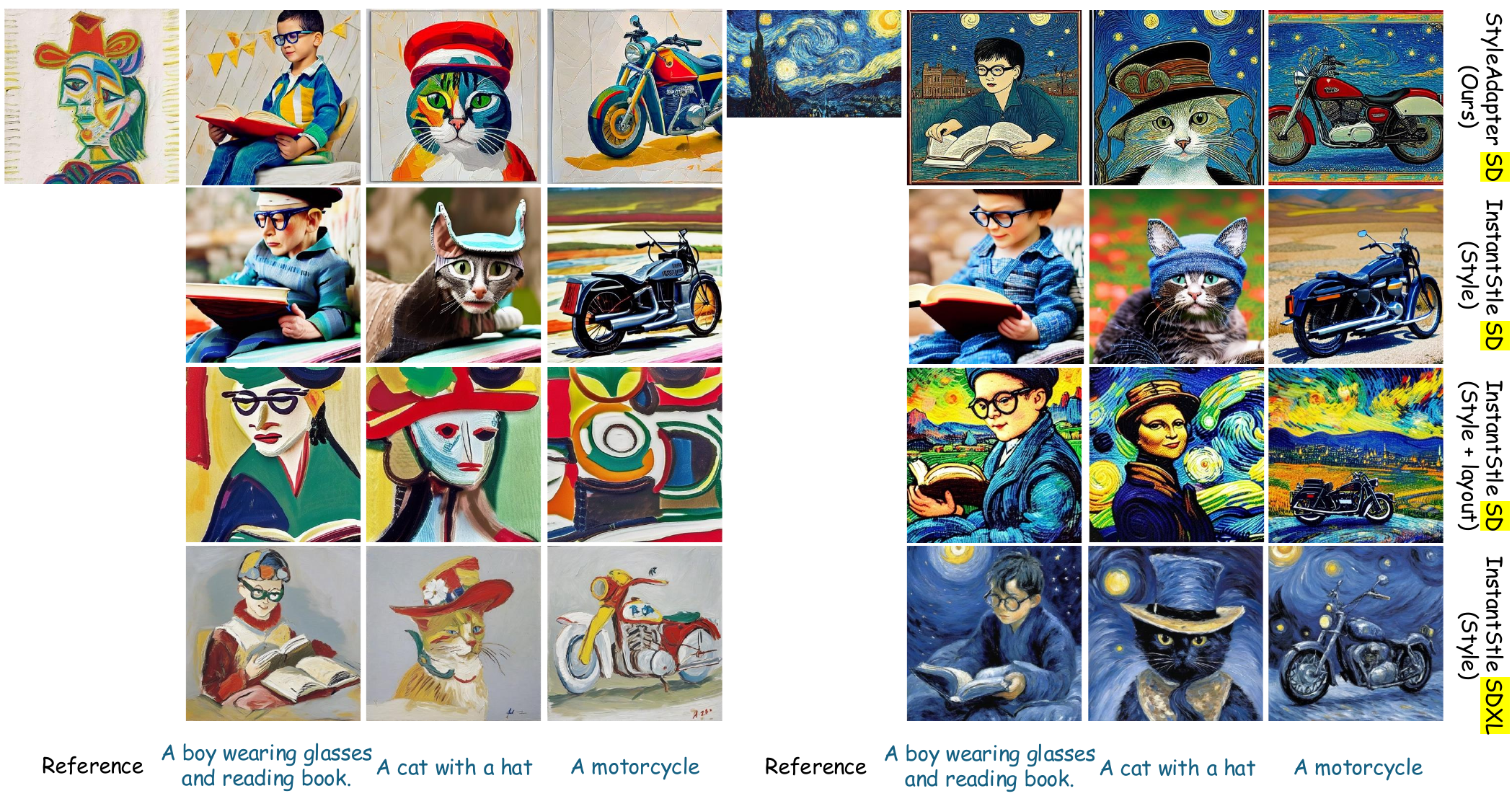}
  \end{minipage}
  \centering
  \caption{
  \redmrtwo{
  \textbf{Comparison with InstantStyle~\cite{Instantstyle}}. Our StyleAdapter, based on SD-v1.5, achieves a more harmonious balance between content and style consistency, thanks to its learnable $\lambda$ and full style integration across all UNet blocks. In contrast, InstantStyle, also based on SD-v1.5, either lacks proper stylization (Style) or misaligns with the text prompt (Style+Layout). Even when InstantStyle is deployed on the more powerful SDXL, which offers better text alignment and generation quality, our StyleAdapter still demonstrates comparable performance.
  }}
  \label{fig:learnable_lamda} 
  \end{figure*}
  
  \subsubsection{Effectiveness of StyleEmb}
  \redmrtwo{
  To evaluate the effectiveness of the learnable embedding $f_m$ in \textbf{StyleEmb}, we remove the learnable embedding and instead combine the features of multiple style references by averaging or concatenating them, using a module identical to StyleEmb except for the learnable embedding. The quantitative and qualitative results, presented in Table~\ref{tab:ab_styleemb} and Fig.~\ref{fig:ab_styleemb}, demonstrate that our carefully designed learnable embedding enhances both content alignment and style consistency. This improvement is attributed to the ability of the learnable embedding to extract similar style features from different references, whereas averaging or concatenating tends to mix all style reference features, negatively impacting the final style transfer and text alignment.
  }
  
  \subsubsection{Effectiveness of Shuffling $E_{pos}$}
  \redmrtwo{
  To suppress the semantic information in the style references, we propose implementing patch-wise shuffling using $E_{pos}$ in the vision model of CLIP~\cite{clip}. Although shuffling can be applied directly to the raw image before CLIP processing, the inherent patchify operation within CLIP further distorts the semantic information of the style image, resulting in inferior stylization in the final output. This is evidenced by a lower style-sim score of 0.8971 compared to 0.9031 achieved through shuffling via $E_{pos}$.
  }
  
  \begin{figure}[ht]
  \centering
  \begin{minipage}[t]{\linewidth}
  \centering
  \includegraphics[width=1\columnwidth]{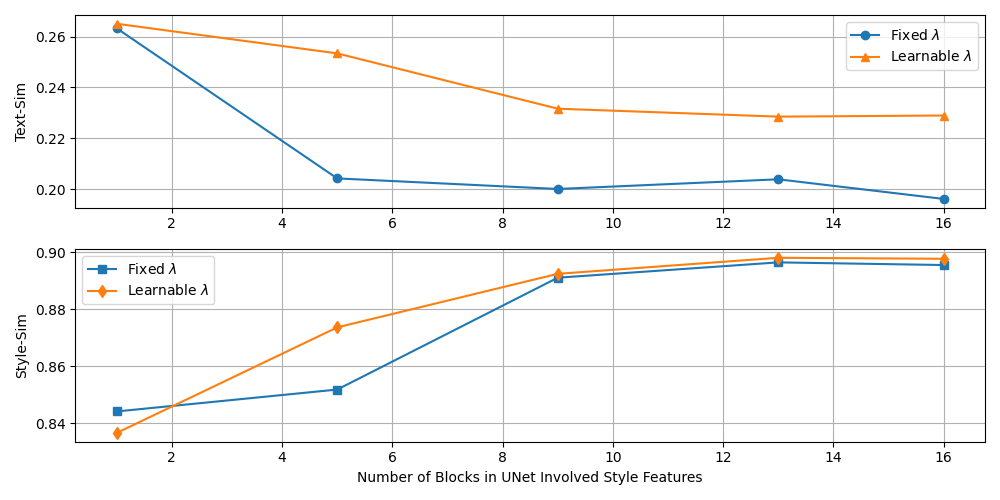}
  \end{minipage}
  \centering
  \caption{
  \redm{\textbf{Leanable $\lambda$.} Progressively expanding style involvement to larger-scale downsampling and upsampling blocks from one middle block in UNet increases the performance of stylization but at the cost of reduced content consistency. However, our StyleAdapter, with a learnable $\lambda$, achieves a better balance in maintaining both content and style consistency.
  }}
  \label{fig:leanable_lamda} 
  \end{figure}
  
  \begin{figure*}[ht]
  \centering
  \begin{minipage}[t]{\linewidth}
  \centering
  \includegraphics[width=1\columnwidth]{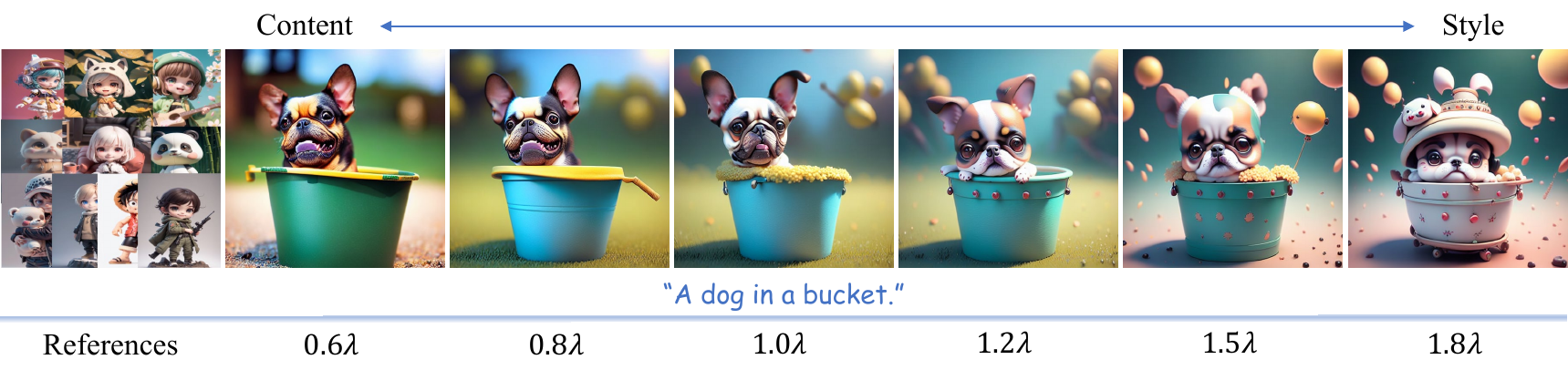}
  \end{minipage}
  \centering
  \caption{
  \textbf{Adaptation of $\lambda$.} By tuning $\lambda$ with an appropriate factor, we can obtain a generated image with a better balance between the content from the prompt and the style from the references. Factors smaller than $1.0$ tend to suppress style features and produce a more natural image, while factors larger than $1.0$ tend to enhance style features.
  }
  \vspace{-10pt}
  \label{fig:adaptive_lamda} 
  \end{figure*}
  
  \begin{table*}[th]
  \caption{\redm{\textbf{Discussion of Efficiency and Model Size.} To accommodate various styles, the model size and training and inference time of our StyleAdapter are relatively larger. However, as the number of style types (N) required to be processed increases, the model size and training time of the other methods also grow proportionally.}}
  \centering
    \begin{tabular}{c|ccccc} 
      \hline
      Methods & InST~\cite{InST} & ProSepct~\cite{Prospect} & TI~\cite{TI} & LoRA~\cite{lora} & \textbf{StyleAdapter} \\
      \hline
      Model Size    & 15M * N & 121M * N & 3.8K * N & 37M * N & 316M \\
      Training Time & $\approx$ 3h * N & $\approx$ 6min * N  & $\approx$ 1h * N & $\approx$ 19min * N & $\approx$ 4days \\
      Inference Time & 4.15s & 5.95s & 4.97s & 5.75s & 8.85s \\
      \hline
    \end{tabular}
    \label{tab:efficiency}
  \end{table*}
  
  \redm{\subsubsection{Learnable and Adaptive $\lambda$}}
  
  \redm{\noindent\textbf{Learnable $\lambda$.}
  In this work, we combine text prompts and style reference images using a learnable $\lambda$ in Eq.~\ref{eq:tpca}, aiming to attain block-wise $\lambda$ values suitable for achieving various aspects of style, such as color, material, brushstrokes, textures, design, atmosphere, and more, which are claimed to be sensitive to different blocks within Unet~\cite{P+}.
  To assess the impact of the learnable $\lambda$ in our StyleAdapter, we conducted an experiment where we fixed $\lambda=1$ during training. The quantitative results, corresponding to $16$ UNet blocks shown in Fig.~\ref{fig:leanable_lamda}, reveal a modest improvement in stylization (as measured by Style-Sim) but a pronounced enhancement in content consistency (measured by Text-Sim).
  Additionally, we explore gradually injecting the style feature across UNet blocks. Starting with only the middle block (where $\lambda$ of other blocks is set to 0), we progressively expanded style involvement to larger-scale downsampling and upsampling blocks until all $16$ blocks incorporated style features. The statistical results in Fig.~\ref{fig:leanable_lamda} demonstrate that increasing the number of blocks with style features enhances stylization effectiveness, but at the cost of content consistency. However, our StyleAdapter with a learnable $\lambda$ achieves a more stable balance between content consistency and stylization performance.
  }
  
  To provide a more intuitive perspective, Fig.~\ref{fig:learnable_lamda} presents the visualized results of our StyleAdapter and \\InstantStyle~\cite{Instantstyle}. InstantStyle is built on Ip-Adapter~\cite{Ip-adapter}, fixing $\lambda=1$ during training, and suppresses the semantics of the style reference by subtracting embeddings and incorporating style information in only a limited number of UNet blocks.
  \redmrtwo{
  The results based on SD-v1.5 are shown in the second and third rows of Fig.~\ref{fig:learnable_lamda}. These demonstrate that InstantStyle, which uses only one (Style) or two (Style+Layout) blocks for style features, either lacks stylization (Style) or fails to align with the text prompt (Style+Layout). In contrast, our StyleAdapter, also built on SD-v1.5, achieves a better balance between content and style consistency due to the learnable $\lambda$ and the involvement of style features in all UNet blocks. Besides, our results implemented with SD-v1.5 are comparable to InstantStyle deployed on SDXL~\cite{podell2023sdxl}, which offers better text alignment and generation quality compared to SD-v1.5.
  }

  \noindent\textbf{Adaptive $\lambda$.}
  After learning, our $\lambda$ is also Adaptive. As shown in Fig.~\ref{fig:adaptive_lamda}, when we scale down $\lambda$ by a factor smaller than $1.0$, the style features from the references fade away gradually, and the generated images become more natural. On the other hand, when we scale up $\lambda$ by a factor larger than $1.0$, the style features in the generated images become more prominent, such as the 3D shape and fantastic appearance. However, the dog also loses its natural look. Therefore, users can customize the generated results according to their preferences by adjusting $\lambda$. The results shown in this paper are obtained with the original $\lambda$ without any scaling factor unless otherwise stated.

  \redm{\subsection{Discussion of Efficiency and Model Size}}
  
  \redm{
  In this section, we discuss the model size, training overhead, and inference time of our StyleAdapter compared to previous related works. Results are in Table~\ref{tab:efficiency}. They are obtained with the default settings of each method and evaluated on a V100-SXM2-32G. Note that N in the table denotes the types of styles to be processed.
  These results reveal that as a unified stylized image generation model that can generalize to various styles without further fine-tuning, StyleAdapter requires a larger model size and longer training and inference time. In comparison, the model size and training time of a single model of the other methods are smaller. However, these models correspond to only one style, and as the number of types of styles (N) increases, their model size and training time grow proportionally. In the real world, there are hundreds and thousands of styles (refer to CIVITAI~\cite{civitai}). Compared to training a model for each style, the advantage of our unified StyleAdapter in terms of training overhead and model size becomes pronounced.
  }

  \subsection{Limitations and Future Work}
  
\begin{figure}[t]
  \centering
  \begin{minipage}[t]{\linewidth}
  \centering
  \includegraphics[width=0.98\columnwidth]{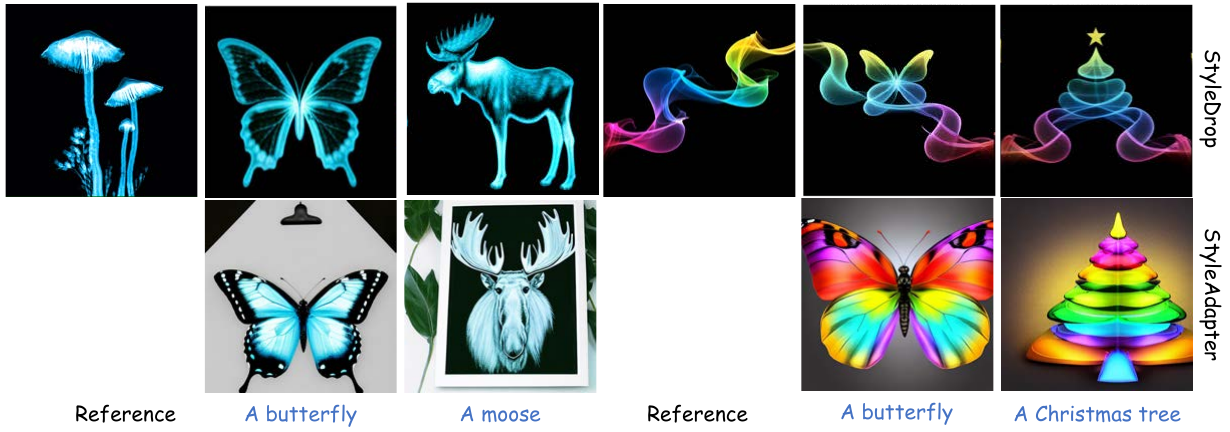}
  \end{minipage}
  \centering
  \caption{\redm{\textbf{Limitation.} As a unified stylized image generation method that does not require per-style fine-tuning, StyleAdapter can capture the color distribution, brushstrokes, and texture from the reference images and apply them to the generated image while maintaining content consistency with the text prompt. However, it cannot fully capture transparent style that seldom appears in the general data used to train our StyleAdapter. We acknowledge that StyleDrop performs better than StyleAdapter in these kinds of styles, benefiting from their relatively labor-intensive training strategy that iteratively fine-tunes the per-style model with human feedback.}}
  \label{fig:limitation} 
  \end{figure}

  \redm{
  As a unified stylized image generation method that does not require per-style fine-tuning, StyleAdapter can be generalized to various styles and performs better in capturing the color distribution, brushstrokes, and texture from the reference images while maintaining the controllability of the text prompt over the generated content, as shown in Fig.~\ref{fig:limitation}. However, StyleAdapter is limited in processing relatively complex styles, such as transparency in the references of Fig.~\ref{fig:limitation}, because there is seldom similar data in the general training data (LAION-AESTHETICS~\cite{schuhmann2022laion}) used for our StyleAdapter. 
  We acknowledge that some per-style training methods perform better in these styles, such as StyleDrop~\cite{sohn2023styledrop}, which benefits from its relatively labor-intensive training strategy that iteratively fine-tunes the per-style model with human feedback. Attaining a more comprehensive training dataset and designing a more robust algorithm to further enhance the generalizability of StyleAdapter is part of our future work.
  }
  

  \section{Conclusion}
  \label{sec:conclusion}
  
  In this paper, we propose StyleAdapter, a unified stylized image generation model capable of producing a variety of stylized images that match both the content of a given prompt and the style of reference images, without the need for per-style fine-tuning. It introduces a two-path cross-attention (TPCA) module to separately process style information and textual prompts, which cooperate with a semantic suppressing vision model (SSVM) to suppress the semantic content of style images. This design is motivated by our in-depth observations and analyses. TPCA ensures the controllability of the prompt over the content of the generated images while SSVM mitigates the negative impact of semantic information in style references, and finally attains high-quality stylized images that conform to both the prompt and style references. 
  
  \noindent\textbf{Data Available.}
  \quad Both the training data and testing data supporting the finding of this work are available at the following URLs: https://laion.ai/blog/laion-5b/, https://civitai.com, https://wall.alphacoders.com, and https://foreverclassicgames.com.
  
  \noindent\textbf{Supplementary Materials.} \quad Additional details of our provided testset, as well as more results and analyses of our proposed StyleAdapter, are comprehensively provided in the supplementary materials. 
  
  \begin{acknowledgements}
    This paper is partially supported by the National Key R\&D Program of China No.2022ZD0161000 and the General Research Fund of Hong Kong No.17200622 and \\17209324. 
    \end{acknowledgements}
  
  \bibliographystyle{spmpsci}      
  \bibliography{reference} 
  
  \end{document}